\documentclass{article}
\usepackage{acra}

\usepackage{float}
\usepackage{graphicx}
\usepackage{amsmath} 
\usepackage{amssymb}  
\usepackage{subcaption}
\usepackage{algorithm}
\usepackage{algpseudocode}
\DeclareMathOperator*{\argmax}{arg\,max}
\DeclareMathOperator*{\argmin}{\arg\!\min}

\title{Learning to Navigate by Growing Deep Networks}
\author{Thushan Ganegedara, Lionel Ott and Fabio Ramos \\ University of Sydney, Australia \\ 
tgan4199@uni.sydney.edu.au and \{lionel.ott,fabio.ramos\}@sydney.edu.au}

\begin{document}

\maketitle

\begin{abstract}
Adaptability is central to autonomy. Intuitively, for high-dimensional learning problems such as navigating based on vision, internal models with higher complexity allow to accurately encode the information available. However, most learning methods rely on models with a fixed structure and complexity. In this paper, we present a self-supervised framework for robots to learn to navigate, without any prior knowledge of the environment, by incrementally building the structure of a deep network as new data becomes available. Our framework captures images from a monocular camera and self labels the images to continuously train and predict actions from a computationally efficient adaptive deep architecture based on Autoencoders (AE), in a self-supervised fashion. The deep architecture, named Reinforced Adaptive Denoising Autoencoders (RA-DAE), uses reinforcement learning to dynamically change the network structure by adding or removing neurons. Experiments were conducted in simulation and real-world indoor and outdoor environments to assess the potential of self-supervised navigation. RA-DAE demonstrates better performance than equivalent non-adaptive deep learning alternatives and can continue to expand its knowledge, trading-off past and present information.
\end{abstract}

\section{Introduction}
Autonomous robot navigation has a broad spectrum of applications, ranging from search and rescue to transportation. Typical approaches to introduce autonomy rely on heuristics or require a predefined set of manually specified rules. Alternative solutions based on machine learning generally require a training set which remains fixed after the learning phase. 
In this paper, we explore alternatives where learning takes place in an online fashion, allowing for adaptability to new circumstances. In this context, self-supervised navigation can be highly advantageous as it allows a robot to train a classifier in real-time, during navigation, not relying on a human driver's knowledge about the environment. This, in turn, allows the robot to explore the environment more effectively without favoring a potential supervisor's bias. Therefore, we  devise a self-supervised technique capable of navigating in an unknown environment through self exploration. 

Vision based autonomous robot navigation is prevalent in many domains such as environmental monitoring~\cite{lee2012vision}, search and rescue~\cite{giustimachine}, off-road driving~\cite{muller2005off}, unmanned aerial vehicle (UAV) maneuvering~\cite{ross2013learning}, reconnaissance etc. Such proliferation of vision-based applications comes at no surprise due to the development of fast and accurate sensors. However, processing sensory data remains challenging, generally requiring feature engineering or large sets of manually labeled data.

\begin{figure*}[t]
\hspace{-0.5cm}
\begin{subfigure}[t]{0.55\textwidth}
 
  \includegraphics[width=0.8\textwidth]{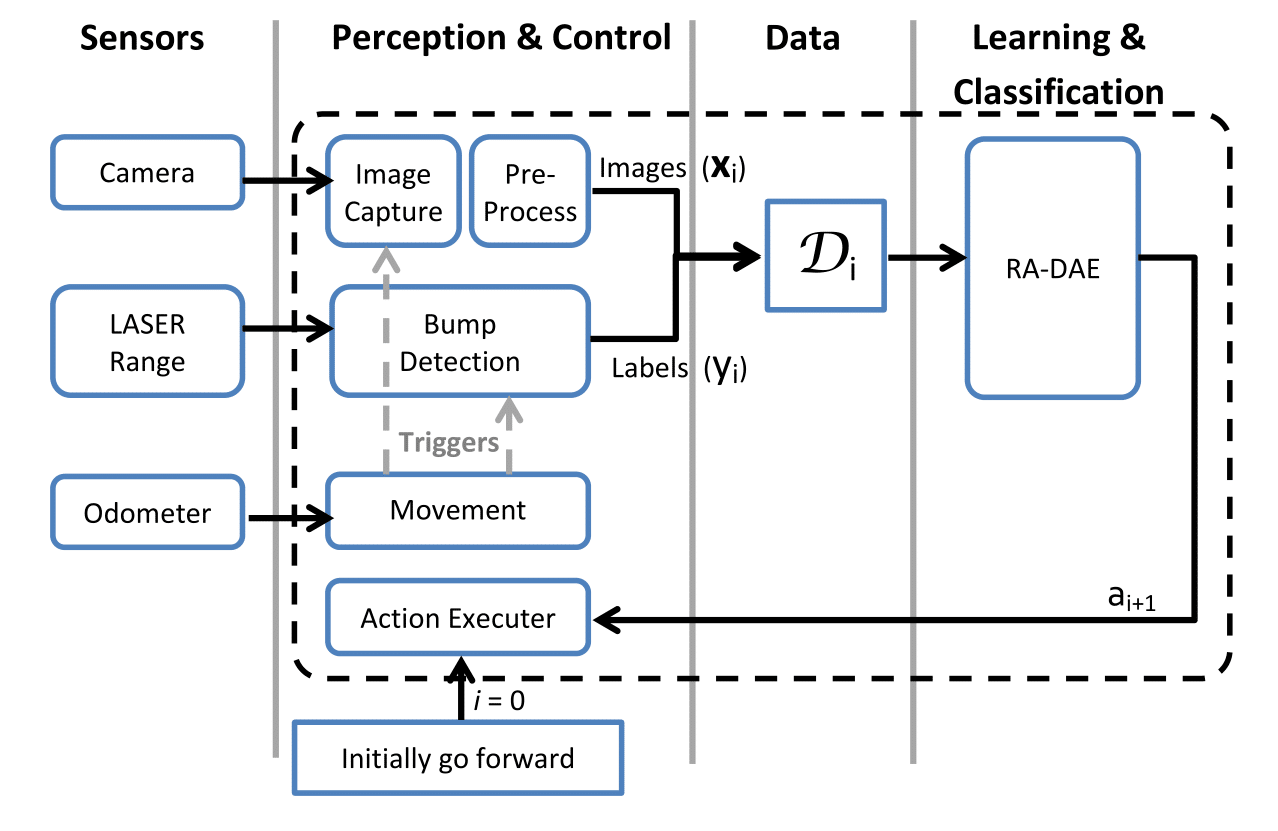}
  \vspace{-0.4cm}
  \centering
  \caption{An illustration of the navigation framework comprising of four major modules; sensing, perception and control, data acquisition, and learning. The LASER Range Finder is solely used to simulate collisions to avoid damages. Initially the robot executes a forward move to collect initial data. This triggers the movement detection module which subsequently triggers both image capture and collision detection modules. This produces $\mathcal{D}^i$, where $\mathbf{x}^i$ represents an image, and $\mathbf{y}^i$ the result of an action, collided or not. Finally, $\mathcal{D}^i$ is fed to RA-DAE that outputs $a^{i+1}$ which is sent to the action executor. The execution of the action triggers the movement module forming a cycle.}
  \label{fig:auto-nav-system}
\end{subfigure}%

\hspace{10cm}
\begin{subfigure}[t]{.42\textwidth}
\vspace{-8.5cm}
\centering
\includegraphics[width=0.8\textwidth,angle=270,trim={3cm 3.5cm 7.9cm 14cm},clip]{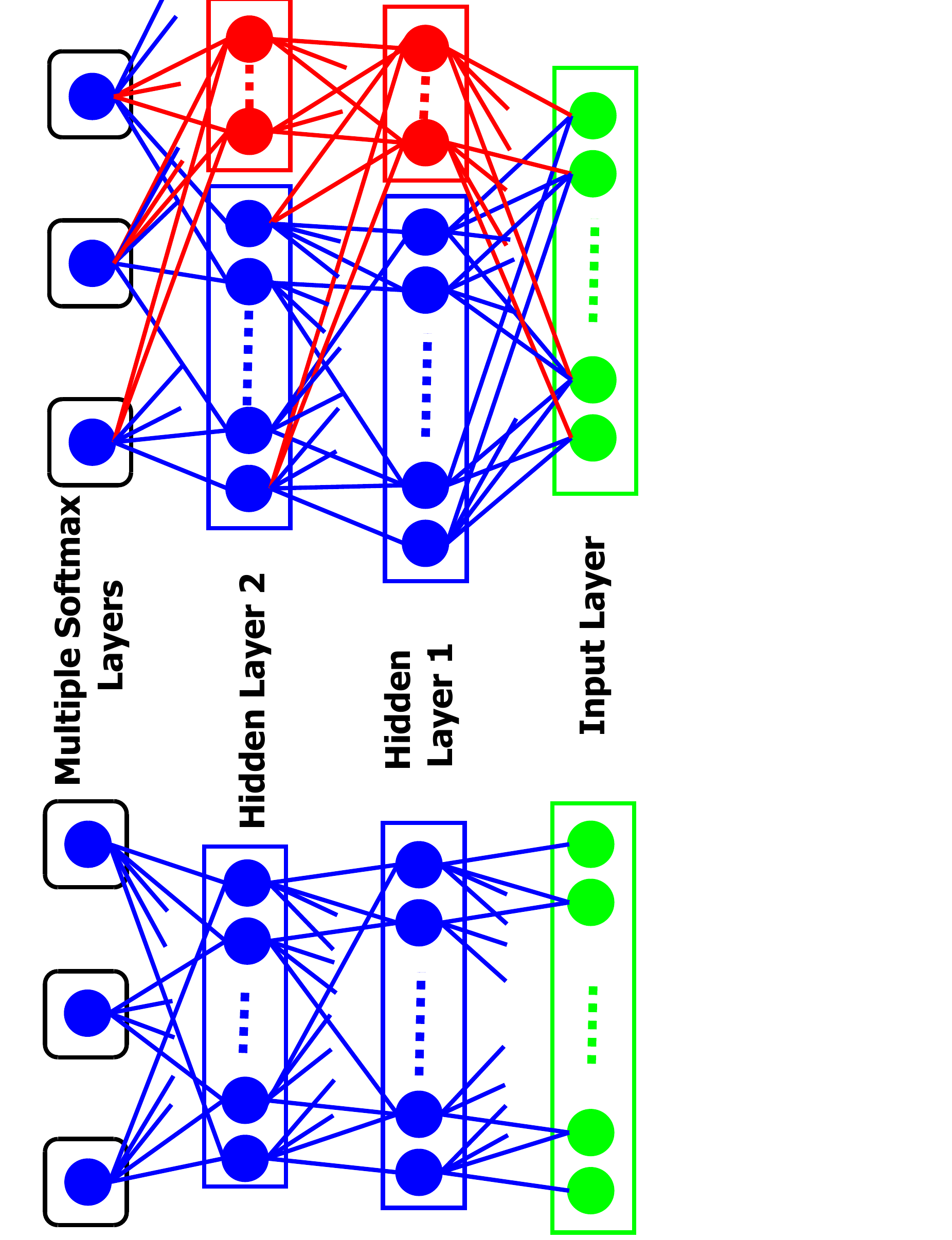}
\vspace{-1.8cm}
\caption{The structure of a two-layer Reinforced Adaptive Denoising Autoencoder (RA-DAE) before (left-side) and after adaptation (right-side). The model uses multiple single node softmax layers (top layer) for each action. The blue lines denote already existed connections and the red lines denote the newly added connections. New connections are introduced while preserving the fully-connected nature. Note that the addition or removal of nodes occurs in a layer-wise fashion (i.e. not simultaneous).}
\label{fig:radae}
\vspace{0.5cm}
\end{subfigure}
\vspace{-0.2cm}
\caption{Block diagram and network structure of RA-DAE.}
\vspace{-0.5cm}
\label{fig:auto_nav_and_radae}
\end{figure*}

We identify several drawbacks of existing approaches that motivate our method. For example, many of them rely on stereo cameras or laser range finders which can be expensive and computationally prohibitive, e.g. computational complexity of the stereo vision systems~\cite{guzel2011vision}. Recent end-to-end learning approaches are based on deep Convolutional Neural Networks (CNNs)~\cite{muller2005off,scoffier2010fully} with a \emph{fixed} structure, leaving many parameters (e.g. number of neurons and layers) to be hand-tuned in order to achieve optimal performance. Furthermore, human supervision is required to provide large training sets, while training is performed offline. This has several disadvantages such as adding significant human bias to a specified task, and sensitivity to changes in the environment, e.g. lighting and weather conditions.

In this paper, we propose a novel approach for a robot to navigate in an unknown environment. Our approach has several appealing characteristics, as it (i) relies on minimal sensory data; a single monocular wide-angle camera, (ii) has an end-to-end, real-time and self-supervised learning process allowing the robot to train and take decisions online (by mapping camera images to actions) without the need for pre-training or pre-labeled data, and (iii) provides a learning algorithm that adapts the structure of a deep network (i.e. number of neurons) on-demand, thus increasing the complexity of the model only as required. For this we devise a Reinforced Adaptive Denoising Autoencoder (RA-DAE) that automatically learns the number of neurons required in the network based on current performance in an online fashion. The interaction of the robot with the environment, i.e. whether it has collided or not into obstacles is used to train the system in a self-supervised manner. The network is initialized with a small number of neurons, growing progressively, as more data becomes available and the complexity of the navigation task increases.

\section{Related Work}

A plethora of approaches have been proposed for vision based navigation over the last decades. Many of the previous work heavily relied on feature engineering. For example scale invariant feature transform (SIFT) \cite{lee2012vision,farag2004detection}, optical flow \cite{lookingbill2007reverse,hyslop2010autonomous} and voxel based~\cite{bagnelllearning,wellington2004online} autonomous navigation techniques can be found in the literature. However, the features learnt by deep learning algorithms have shown to perform better than hand-crafted features.



Recently, deep learning techniques have been adopted in a multitude of robotics applications. Deep learning techniques~\cite{lecun2015deep} are renowned for their ability to jointly perform feature extraction and classification using raw data. Furthermore, deep networks have demonstrated unprecedented performance in certain cognitive tasks such as traffic sign recognition~\cite{cirecsan2012multi} and pedestrian detection~\cite{sermanet2013pedestrian}. Inspired by the state-of-the-art performance of deep neural networks, especially CNNs have been leveraged successfully for visual navigation for autonomous robots~\cite{muller2005off,giustimachine}. However, these approaches still mainly rely on human supervision or high-quality sensory information produced by several equipment such as Stereo Cameras or Light Detection And Ranging (LIDAR) posing limitations for adapting such methods for real-time navigation.

Furthermore, the idea of adapting the structure of neural networks has been around for a long time. One popular approach was to use genetic algorithm to evolve the structure of the network guided by a fitness function~\cite{stanley2002evolving}. This method has been successfully used in various robotics application~\cite{de2009method,stanley2004competitive}. However for high-dimensional raw sensory inputs (i.e. images) and complex multi-layer networks, it becomes computationally infeasible due to the large number of possible combinations.

Finally, ~\cite{courbon2009autonomous} proposes navigation techniques that rely only on cheap and low power devices such as a single monocular camera. Despite their performance, these techniques require a human guiding the robot through the environment during training. This could be costly for unknown or human-inaccessible terrains. Moreover, ~\cite{courbon2009autonomous} relies on persisted visual memory (images) for navigation, which does not scale well for navigating large spaces.

\section{Overview}

Figure~\ref{fig:auto-nav-system} depicts the high-level architecture of our framework. Our method is an end-to-end learning process which converts images captured by a monocular camera into navigation commands for the robot. The approach comprises several vital components that together learn in real-time. Our approach consists of the following steps. During the execution of each action, tuples of images of what the robot perceives and labels are collected. Each label is an integer indicating whether the robot has collided or not during the execution of an action, i.e. 0 or 1. We define the actions of the robot as discrete movements. The robot can turn left ($L$), go straight ($S$) or turn right ($R$). Each of these movements move the robot by a fixed $\delta$ (step-size) distance in the corresponding direction. Then several pre-processing operations are executed on the collected images such as normalization. Next the accumulated collection of tuples of images and labels are fed to the learning algorithm. The learning model trained on the received data, converts the images into actions (i.e. movements). This procedure is repeated for each action and associated tuples of images and labels.

As the data being collected grows, we need an online mechanism to quickly adapt to new information. 
Reinforced Adaptive Denoising Autoencoder (RA-DAE)~\cite{ganegedara2016online} is a deep learning technique that uses reinforcement learning to dynamically adapt the structure of a deep network as the data distribution changes. Such adaptations include adding neurons, merging neurons, and fine tuning. Figure ~\ref{fig:radae} illustrates the resulting adapted network after adding neurons. RA-DAE leverages Q-Learning~\cite{sutton1998reinforcement}; a reinforcement learning (RL) technique, to find the best adaptation settings based on the errors made during the training phase. 

The main motivation for our approach is that the vanilla deep network techniques do not possess the ability to adapt their structure to compensate for changes in data distribution. Such changes in data distribution, known as \emph{covariate shift} can cause \emph{catastrophic forgetting} in the networks. RA-DAE not only has the ability to adapt the structure, but also strives towards finding best adaptation strategy (i.e. add neurons, remove neurons or no change) for the perceived changes in the data distribution. With such capabilities, RA-DAE creates an opportunity for deep networks to be used for robotics applications by cutting down on the training time and the prediction time as well. This is enabled by RA-DAE's ability to start with a small neural network and grow the network by small incremental steps as needed.  

\section{Background}
This section provides a brief description of Stacked Denoising Autoencoders as the basic model used by RA-DAE. We begin by defining notation. 

\noindent \textbf{Notation:} Let us assume we have a data stream $\mathcal{D}=\{(\mathbf{x}^1,\mathbf{y}^1),(\mathbf{x}^2,\mathbf{y}^2),(\mathbf{x}^3,\mathbf{y}^3),\ldots\}$ where $\mathbf{x}^i=\{x^{i,1},x^{i,2},\ldots,x^{i,d}\}$, $d$ is the dimensionality of a single input and $\mathbf{y^i} \in \{0,1\}^K$ such that if $y^{i,j}$ are the elements of $\mathbf{y}^i$ then $\sum_j{y^{i,j}}=1$. The $n^{th}$ batch of data in $\mathcal{D}$ is written as $\mathcal{D}^n=\{\{\mathbf{x}^{n-p},\mathbf{y}^{n-p}\},\ldots,\{\mathbf{x}^n,\mathbf{y}^n\}\}$ where $p$ is the batch size.\\ 
\vspace{-0.2cm}
\subsection{Autoencoder}
The autoencoder aims to map input data (dimensionality $d$) into a latent feature space (dimensionality $H$) with a series of nonlinear transformations $h_{W,b}(\mathbf{x})=sig(W\mathbf{x}+b)$, and reconstruct the original input with $\mathbf{\hat{x}}=sig(W^T \times h_{W,b}(\mathbf{x})+b')$ from the latent feature space, where $sig(s) = \frac{1}{1+\exp{-s}}$ and $W \in {\rm I\!R}^{H\times d}$, $b\in{\rm I\!R}^{H\times 1}$ and $b'\in{\rm I\!R}^{d\times 1}$ are the parameters of the autoencoder. This is achieved by optimizing the parameters of the network with respect to the generative (i.e. reconstruction) error $L_{gen}(\mathbf{x}^i,\mathbf{\hat{x}}^i)=\sum_{j=1}^{d}x^{i,j}\text{log}(\hat{x}^{i,j}) + (1-x^{i,j})\text{log}(1-\hat{x}^{i,j})$ $\forall \mathbf{x}^i$ where $\mathbf{x}^i$ is the input and $\mathbf{\hat{x}}^i$ is the reconstructed input. Notice that the learning in an autoencoder is unsupervised.\\
\vspace{-0.2cm}
\subsection{Stacked Autoencoders}
By stacking $J(>1)$ autoencoders vertically, and topping it with a classification layer e.g. softmax, the construction can be leveraged to solve a supervised classification task. Such networks are called stacked autoencoders (SAE)~\cite{vincent2010stacked}. In the training process of SAE, the predicted label, $\mathbf{\hat{y}} = \text{softmax}(W^{out}h_{W,b}^J(\mathbf{x})+b^{out})$ is calculated for input $\mathbf{x}$ where $h_{W,b}^J(\mathbf{x})$ is the output of the $J^{th}$ autoencoder and softmax($a_k$) = $\frac{\exp(a_k)}{\sum_{k'}\exp(a_{k'})}$ where $a\in[0,1]^K$. Then all the parameters ($W^1,\ldots,W^J$,$b^1,\ldots,b^J$,$b'^1,\ldots,b'^J$,$W^{out},b^{out}$ and $b'^{out}$) are optimized with respect to two error measures; the generative error $L_{gen}(\mathbf{x},\mathbf{\hat{x}})$ and the discriminative (i.e. classification) error $L_{disc}(\mathbf{y},\mathbf{\hat{y}})$ $\forall \{\mathbf{x},\mathbf{y}\} \in \mathcal{D}$ where $L_{disc}(\mathbf{y},\mathbf{\hat{y}}) = \sum_{j=1}^K(y^j\text{log}\hat{y}^j + (1-y^j)\text{log}(1-\hat{y}^j))$ and $\{W^i,b^i,b'^i\}$ are the parameters of the $i^{th}$ autoencoder. 
\newline
\vspace{-0.5cm}
\subsection{Stacked Denoising Autoencoders}
Stacked Denoising Autoencoders~\cite{vincent2010stacked} is an improvement over SAE that attempts to reconstruct inputs based on corrupted versions of the inputs leading to more robust features. A common way of achieving this is to mask the input with a binomial distribution with probability $p$ where $1-p$ is the corruption level. This procedure improves the generalization properties of stacked autoencoders by acting as regularization.   

\section{RA-DAE}
RA-DAE employs a similar approach to SDAE to learn the network from training data. However, RA-DAE adopts a novel approach as it leverages reinforcement learning to make dynamic adaptations to the structure of the network as the observed data distribution changes. The problem of adapting the network over time is formulated as a Markov Decision Process (MDP) with the state space (S), action space (A) and a reward function ($r^n$) defined as follows. 

\subsubsection{State Space}
The state space is defined as 
\begin{align} \label{eq:cont-states}
S=\{\mathcal{\tilde{L}}_{g}^{n}(m), \mathcal{\tilde{L}}_{c}^n(m), \nu_1^n \} \in {\rm I\!R}^3
\end{align}

\noindent where the moving exponential average ($\mathcal{\tilde{L}}$) is defined as $\mathcal{\tilde{L}}^n(m) = \alpha \text{L}^n + (1-\alpha) \mathcal{\tilde{L}}^{n-1}(m-1)$, $n \geq m$ and $m$ is a predefined constant. $\mathcal{\tilde{L}}_{g}^n$ and $\mathcal{\tilde{L}}_{c}^n$ denote $\mathcal{\tilde{L}}^n$ w.r.t. L$_{g}^n$ and L$_{c}^n$, where L$_g^n$ and L$_c^n$ are the average generative and discriminative errors for the $n_{th}$ batch of data, respectively, and  $\nu_l^n = \frac{\text{Node Count}_{current}}{\text{Node Count}_{initial}}$ for the $l^{th}$ hidden layer. $\mathcal{\tilde{L}}$ is defined in terms of recursive decay to respond rapidly to immediate changes. 

\subsubsection{Action Space}
The action space is defined as,
\begin{equation} \label{eq:actions}
A = \{Pool, Increment(\Delta), Merge(\Delta)\},
\end{equation}
where $\Delta$ is a pre-defined constant representing the number of nodes. We define two pools of data B$_{ft}$ and B$_r$ to be utilized by the actions in Equation \ref{eq:actions}. B$_{ft}$ is composed of the $\tau$ most recent incorrectly classified batches of data, as detailed in Equation \ref{eq:b_ft}. B$_r$ contains the $\tau$ most recently observed batches, and $\tau$ is a predefined constant.

Increment($\Delta$) adds $\Delta$ new nodes and greedily initializes them using pool B$_r$. The Merge($\Delta$) operation is performed by merging the 2$\Delta$ closest pairs of nodes into $\Delta$ nodes. The Pool operation trains the network with B$_{ft}$ given its previous parametrization. 

\subsubsection{Reward Function}
The reward function is defined as follows,
\begin{equation}
	\label{eq:rn_pean}
  	r^n= \begin{cases}
		g^n - |U - \nu_1^n| & \text{if } \nu_1^n < V_1 \text{ or } \nu_1^n > V_2 \\ 
		g^n & \text{otherwise}
    \end{cases},
\end{equation}

\noindent where $g^n =(1-(\text{L}_{c}^n-\text{L}_{c}^{n-1}))\times(1-\text{L}_{c}^n)$ and $U, V_1$ and $V_2$ are predefined thresholds penalizing the network if it grows too large or small.

With the definition of $S$, $A$ and $r^n$, Q-Learning~\cite{sutton1998reinforcement} is employed to learn a desirable policy, i.e. a function that defines which action to take in a given state, to control the structural changes. Q-Learning is a reinforcement technique that learns policies without relying on a deterministic model of the environment. This is a desirable property to have as the environment of our MDP is complex and only partially-observable. The desired policy is learned by updating an utility function $Q(s,a)$ which quantifies the reward for executing action $a$ in state $s$.

In RA-DAE, Q-Learning is used in the following way. For the $n^{th}$ iteration, with data batch $\mathcal{D}^n$,
\begin{enumerate}
\item Until adequate samples are collected, i.e. $n \leq \eta_1$, train the network with B$_r$.
\item With adequate samples collected, i.e. $n>\eta_1$, start calculating Q-values for each state-action pair observed $\{s^n,a^n\}$, where $s^n \in S$, and $a^n \in A$.
\item When $\eta_1 < n \leq \eta_2$, uniformly perform actions from $A=\{$\emph{Increment}, \emph{Merge}, \emph{Pool}$\}$ to develop a descent value estimate for all actions in $A$.
\item With an accurate estimation of $Q$, i.e. $n>\eta_2$, the action $a'$ is selected by $a'=\argmax _{a'}(Q(s^n,a'))$ with a controlled amount of exploration ($\epsilon$-greedy).
\item Execute action $a'\in A$, train the network with $\mathcal{D}^n$ and finally calculate the new state, $s^{n+1}$, and the reward $r^{n}$.
\item Update the value (utility) $Q(s,a)$ as, 
$Q^{(t+1)}(s^{n-1},a^{n-1}) = (1-\alpha) Q^{t}(s^{n-1},a^{n-1})
+ \alpha \times q$,

where $q=r^n + \gamma \times \max _{a'}(Q^{t}(s^n,a'))$,
and $\eta_1$, $\eta_2$, the learning rate $\alpha$, and the discount rate $\gamma$ are predefined constants. 
\end{enumerate}

\section{Self-Supervised Navigation}

The objective of this paper is to introduce a real-time self-supervised navigation mechanism that only relies on vision. We use SDAEs to learn the optimal navigation action given the current perception of the robot. Since the learning is performed in real time it is desirable to explore model complexity performance trade-offs to make the learning efficient. This is achieved by using an adaptive variant of SDAEs known as RA-DAEs. Ideally, RA-DAE should increase the complexity of the model as new parts of an environment is being explored and either reduce or keep constant when previously seen parts of the environment are encountered.

In the following we will describe the components of our method for self-supervised navigation using RA-DAE and how they work together in more detail. As the method uses a reinforcement learning framework we consider that each movement of the robot is an episode denoted by $E^i$ with $i = 0, \dots, N$ where $N$ is the number of episodes in a single experiment. The action taken in episode $i$ is denoted by $a^i \in A$ where $A=\{1,\ldots,K\}$ denotes the set of $K$ discrete actions available which each are linked to their individual softmax layer.

During the execution of action $a^{i+1}$ in episode $E^i$ the robot collects a set of images and self-supervised labels. This forms the data set $\mathcal{D}^i=\{(\mathbf{x}^i_1,y^i_1),(\mathbf{x}^i_2,y^i_2),(\mathbf{x}^i_3,y^i_3),\ldots\}$ where $\mathbf{x}^i_j$ represents the pixels in a single image with the associated label $y^i_j \in \{0, 1\}$ denoting if a collision occurred during the execution of action $a^{i+1}$. Using this data we train RA-DAE$_{a^{i+1}}$ by combining the softmax layer for action $a^{i+1}$ and the shared hidden layers, as shown in Figure~\ref{fig:radae}.

When deciding which action $a^\prime$ to execute next we query the RA-DAE$_{a^{i+1}}$ model to obtain the probability of executing action $a^\prime$, i.e. $P_{a^{a+1}}(a^\prime) = P(a^{i+1}=a^\prime \mid \mathcal D^i, \theta_{a^{i+1}})$, where $\theta_{a^{i+1}}$ are the latent variables, i.e. weights, of RA-DAE$_{a^{i+1}}$. With this we can obtain the probability of choosing each action as $\mathbf b^{i+1} = \{P_{a^{i+1}}(a^\prime)\} \forall a^\prime \in A$.




Putting the training and querying parts together into an end-to-end process as illustrated in Figure~\ref{fig:auto-nav-system} which shows the different components of our framework. Initially the system executes action $a^0 = S$, i.e. go straight for $\delta$ meters. Then the next action is selected by computing the probability of each action $a^\prime$ as $\{P_{a^{i+1}}(a^\prime)\} \forall a^\prime \in A$ and evaluating the following action selection function:
\vspace{-.1cm}
\begin{equation}
	\label{eq:action_selection}
	a^{i+1}=\begin{cases}
		\text{random } & \text{if } P_{a^{i+1}}(\hat{a})<\mu_1 \text{ or }\\ 	
        	& P_{a^{i+1}}(\hat{a})>\mu_2 \forall \hat{a} \in A\\
		\argmin_{\hat{a}}(A^\prime) & \text{otherwise}
\end{cases},
\vspace{-.1cm}
\end{equation}

\noindent where $a^\prime \in A^\prime \text{ if } \mu_1 \leq P_{a^{i+1}}(a^\prime) \leq \mu_2\ \forall a\in A$ with $\mu_1$ and $\mu_2$ as predefined constants. The action $a^{i+1}$ selected in this manner is then executed in episode $E^{i+1}$ which yield $\mathcal D^i$ which allows us to train RA-DAE$_{a^{i+1}}$.  Next, RA-DAE$_{a^{i+1}}$ is trained on $\mathcal{D}^{i}$ which contains the observations the robot made while executing $a_{i+1}$ and the labels $\mathbf y^i$ attached to those, i.e. $y^i_j = 0 \quad \forall y^{i}_j \in \mathbf y^{i}$ in case of a collision. As the labels are obtained in a self-supervised manner by the robot this procedure allows it to improve the models in a self-supervised way. This process of picking the next action and improving the model based on the collected observations for that action is repeated until termination, an overview of the algorithm is given in Algorithm~\ref{algo:navigation}.

The algorithm starts by executing its initial action, i.e. move forward. While the robot moves, images are stored. Once the motion terminates the algorithm checks whether or not a collision occurred. If the robot collided he reverses back to the last safe position and trains the RA-DAE model of the executed action with the stored data. The same happens when no collision was detected, with the difference that the robot stays in its current position and saves it as the last known safe position. Once this is done the next action to execute is selected by evaluation the RA-DAE models for each action. The selected action is then executed.

\begin{algorithm}[bt]
\caption{Navigation algorithm}
\label{algo:navigation}
\begin{algorithmic}
\Procedure{Navigate()}{}
\State \textbf{define} : lastSafePos - Last non-collided position
\State $i=0$
\State lastSafePos = current robot position
\State Execute action $a^i = S$
\While {notTerminated}
\While {moving}
\State Accumulate $\mathbf{x^i_j} \hspace{0.1cm}\text{where}\hspace{0.1cm} j=1,2,\ldots$
\State Check collision
\EndWhile
\If {$i>0$ and collision}
\State Reverse to lastSafePos
\State $\mathcal{D}^{i}=\{(\mathbf{x}^{i}_j,y^{i}_j)\}  \hspace{0.1cm}\text{where}\hspace{0.1cm} y^{i}_j=0 \hspace{0.1cm} \forall j$
\State Train RA-DAE$_{a^{i+1}}$ with $\mathcal{D}^{i}$
\EndIf
\If {$i>0$ and not collision}
\State $\mathcal{D}^{i}=\{(\mathbf{x}^{i}_j,y^{i}_j)\}  \hspace{0.1cm}\text{where}\hspace{0.1cm} y^{i}_j=1 \hspace{0.1cm} \forall j$
\State Train RA-DAE$_{a^{i+1}}$ with $\mathcal{D}^{i}$
\State lastSafePos = current robot position
\EndIf

\For {$\forall a \in A$}
\State Calculate $P_{a+i}(a)$ with RA-DAE$_a$
\EndFor
\State $i = i+1$
\State Calculate and Execute $a^{i+1}$ (Equation~\ref{eq:action_selection})
\EndWhile
\EndProcedure
\end{algorithmic}
\end{algorithm}

%
%

Finally, we discuss the two key modifications introduced to RA-DAE to make the algorithm more applicable to navigation tasks. First, we use $K$ single node softmax layers for classification, i.e. one layer for each action (Figure~\ref{fig:radae}). In contrast to the alternative of a single softmax layer with $K$ nodes our approach allows multiple actions to be valid for the same data by imposing more independence between the actions. Second, RA-DAE uses two pools of data: $B_{r}$ to train the newly added neurons and $B_{ft}$ to fine-tune the whole network, the latter was modified as follows:
\begin{equation}
	\label{eq:b_ft}
	B_{ft} = \begin{cases}
		\mathcal D^i \cup B_{ft}  \text{ if } y^{i}_j=1 \forall y^{i}_j \in
        	\mathcal D^i \land 
              y^{i-1}_j = 0 \forall y^{i-1}_j \in
            \mathcal D^{i-1} \\
		\mathcal D^i \cup B_{ft}  \text{ if } y^{i}_j=0 \forall y^{i}_j \in
        	\mathcal{D}^{i} \\
		B_{ft} - \mathcal D^i  \text{ if } |B_{ft}| > \tau 
        	\argmin_{i^\prime}(
        	\forall \mathcal D^{i^\prime} \in B_{ft})  \\
		B_{ft}  \text{ otherwise}
\end{cases} 
\end{equation}

The argument behind the modifications is as follows. As $B_{ft}$ is employed to train the whole network, we fill $B_{ft}$ with the instances our algorithm misclassified. As such, $B_{ft}$ collects data that depicts a wrong action executed or the corresponding correct action.

\section{Experimental Results}

\subsection{Overview and Setup}
\label{sec_overview}
\begin{figure}[t]

\hspace{.5cm}
\begin{subfigure}{.22\textwidth}
\includegraphics[width=0.8\linewidth]{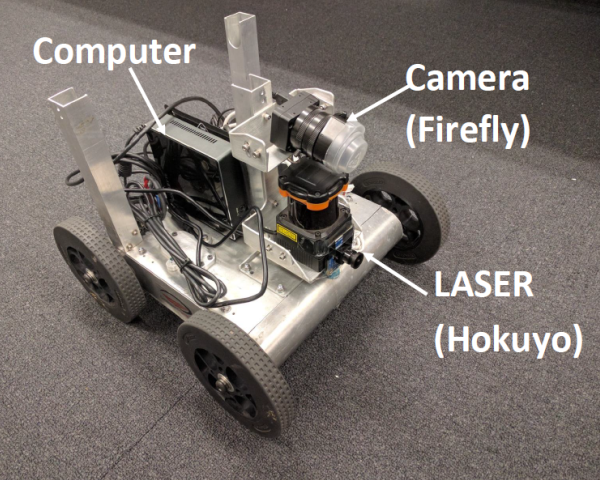}
\centering
\caption{The robot}
\label{fig:wombot}
\end{subfigure}
\hspace{-.5cm}
\begin{subfigure}{.22\textwidth}
  \centering
  \includegraphics[width=.8\linewidth]{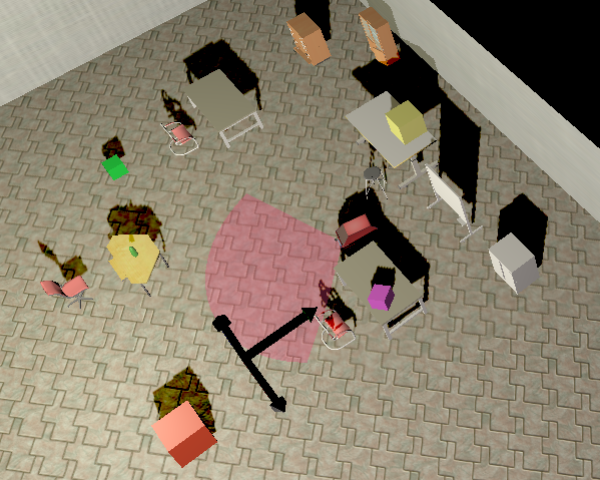}
  \caption{Simulation}
  \label{fig:sim1}
\end{subfigure}%

\hspace{.5cm}
\begin{subfigure}{.22\textwidth}  
  \centering
  \includegraphics[width=.8\linewidth]{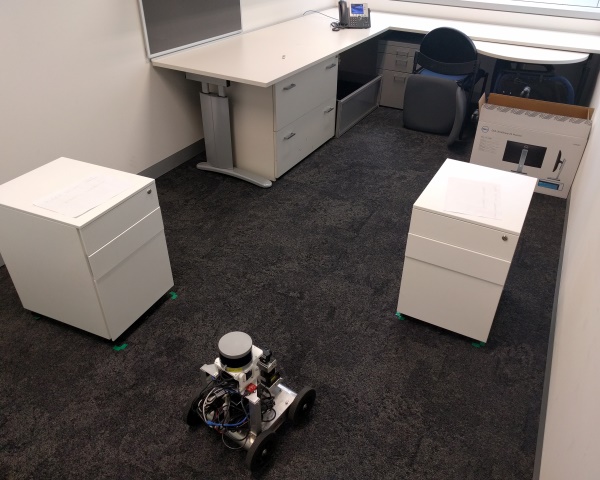}
  \caption{Office}
  \label{fig:loc_office}
\end{subfigure}
\hspace{-.5cm}
\begin{subfigure}{.22\textwidth}
  \includegraphics[width=.8\linewidth]{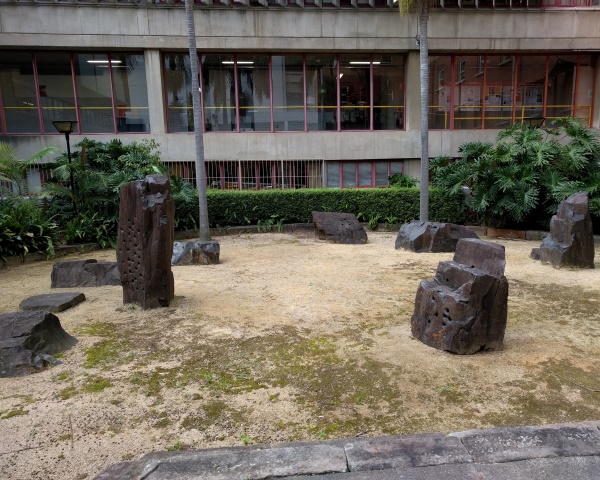}
  \centering
  \caption{Outdoor}
  \label{fig:loc_outdoor}
\end{subfigure}
\vspace{-0.3cm}
\caption{Environments and the robot.}
\vspace{-0.5cm}
\label{fig:env}
\end{figure}

Several experiments were conducted to assess the performance of our approach. The experiments were done in simulation and using a real robot. We used Morse\footnote{https://www.openrobots.org/wiki/morse/} as the simulation framework and an indoor environment already available in the framework. An office and an outdoor area were used as the real-world environments. Our robot (Figure~\ref{fig:env}b) is equipped with a Firefly MV camera producing 640x480 RGB images at 30Hz, a 40Hz Hokuyo laser mounted in front of the robot and an onboard Intel i7-4500U 1.80GHz. The laser scanner is used to detect imminent collisions and avoid damaging the environment, essentially acting as a bump sensor and not providing any range information.  

The approach (RA-DAE) was tested against a standard Stacked Denoising Autoencoder (SDAE) and a Logistic Regression Classifier (LR). For fairness, we introduced multiple single node softmax layers (i.e. layer per action) for both SDAE and LR, and a pooling step for SDAE similar to RA-DAE~\cite{ganegedara2016online} which trains the model on previously seen data.

The following settings were used for all the experiments. Total number of episodes, N=500 for the simulation, and N=400 for the real-world respectively. Simulation experiments were run on a NVidia Tesla K40c while the real-world experiments were only using the onboard computer of the robot.  Theano~\cite{bastien2012theano} was used for the implementations. The parameter $\delta$ (distance traveled before taking an action) was set to $1m$. $\mu_1$ and $\mu_2$ for the action selection algorithm were selected as 0.45 and 0.95 respectively for all algorithms. For all experiments we used a batch size of 5. A smaller batch size was important as the data was collected in real-time. The corruption level (0.15), activation function ({\em sigmoid}) and the learning rates  for RA-DAE (0.01), SDAE (0.05) and LR (0.001) where chosen with a coarse grid search. Different learning rates are required as structural complexities were different for different algorithms. For example, SDAE failed to perform with low learning rates due to the complexity of the network (i.e. large number of weights). RA-DAE and SDAE were initialized with three layers having 64, 48 and 32 neurons, and 256, 196 and 128 neurons, respectively. For RA-DAE, $m$ and $\tau$ were set to 15 and 10000 respectively as a compromise between the memory requirement and performance. $\eta_1$ and $\eta_2$ were set to 5 and 30 in order to provide adequate time for Q-Learning algorithm to explore the action space before predicting actions based on the value function. $\Delta$ was set to 5 to achieve a consistent and smaller growth rate of the network over time due to the limited amount of data possessed. Finally, no regularization was employed except for denoising. We tested \emph{dropout}~\cite{srivastava2014dropout}, however, dropout failed in all the experiments, as such stochasticity disrupts the incremental nature of RA-DAE. 

\begin{table}[t]
\caption{Percentage of collisions and time consumption w.r.t the number of hidden layers of RA-DAE. The two tables denote the results obtained for two distinct starting locations in the simulated map, Fig~\ref{fig:env}a. $L_{NW}$ and $L_{W}$ are the average percentage of the count of collisions and the average of false-positive probabilities of the collisions in the last 250 episodes. The time consumption denotes the average training and prediction time per episode respectively. It can be seen that there is a clear advantage in increasing the number of hidden layers.}
\label{tbl:hidden_layers_1}
\centering
\begin{tabular}{|c|c|c|c|}
\hline
Hidden & \multicolumn{2}{|c|}{Average collision percentage} & Training\\\cline{2-3}
 Layers & $L_{NW}$ & $L_{W}$ & Time (s)\\
 \hline
1 & 27.6$\pm$6.38\%& 16.78$\pm$4.44\% & 0.307\\
3 & \textbf{21.6$\pm$5.64\%} & \textbf{15.09$\pm$3.44\%} & 0.497\\
\hline
\end{tabular}

\vspace{.1cm}

\begin{tabular}{|c|c|c|c|}
\hline
Hidden & \multicolumn{2}{|c|}{Average collision percentage} & Training\\\cline{2-3}
 Layers & $L_{NW}$ & $L_{W}$ & Time (s)\\
 \hline
1 & 31$\pm$11.40\%& 17.11$\pm$4.91\% & 0.326\\
3 & \textbf{26.6$\pm$10.54\%} & \textbf{15.64$\pm$4.61\%} & 0.441\\
\hline
\end{tabular}
\vspace{-0.5cm}
\end{table}

\subsection{Preprocessing}
Following the data acquisition, several low-cost pre-processing steps are performed on the captured images to make the learning more effective. Since the frame rate of the camera (30Hz) and the frequency of the laser (40Hz) are too high and mismatched, they are downsampled to 10Hz. This operation enables us to have a label corresponding to each image captured image. The images are then preprocessed as follows. First the images are resized to 128x96 and cropped vertically (19 pixels from each side) to produce 128x58 images. Next the images are converted to \emph{grayscale} from RGB space and normalized to the range of $[0,1]$. Downsampling the images is important to make the computations feasible in real-time. Finally, the mean is subtracted from the images to produce zero-mean inputs. 

\subsection{Performance Metric}
We define the performance metrics as functions of the number of collisions occurred. The definition of accuracy can be difficult to discern for navigation tasks. For example, a standard error measure such as squared loss does not suit our approach as the labels are a mere reflection of the correctness of an action taken and does not possess information about the validity of other actions at a given time. Therefore, we calculate the non-weighted ($L_{NW}$) and weighted ($L_W$) count of collisions for a given time window to measure the performance. For the time frame $E^{i-M:i}$, where $E^{i-M:i}$ is composed of episodes $\{E^{i-M},\ldots,E^{i-1}\}$, we define, $L_{NW}^{i-M:i}$ as the number of collisions occurred during $E^{i-M:i}$ and $L_{W}^{i-M:i}$ is the weighted number of collisions with weights equal to the probability of executing the action leading to the collision. We used $M = 25$ and $i = \{0, 25, 50, \dots, N\}$.

\subsection{Evaluating the Effect of the Number of Layers}
In order to assess the effect of the number of hidden layers on the performance, we tested a one layer and a three layer RA-DAE in the simulated environment. Table~\ref{tbl:hidden_layers_1} shows the average percentage of collisions per 25 episodes in the last 250 episodes ($L_{NW}$ and $L_{W}$) of a total of 500 episodes as well as the training and prediction time per episode. $L_{NW}$ and $L_{W}$ are calculated by taking the average of $L_{NW}^{i-25:i}$ and $L_{W}^{i-25:i}$ where $i=\{275,\ldots,500\}$ and converting them to percentages.
It can be seen that deeper models deliver better performance. 

\begin{figure}[t]
\centering
\includegraphics[width=0.49\textwidth]{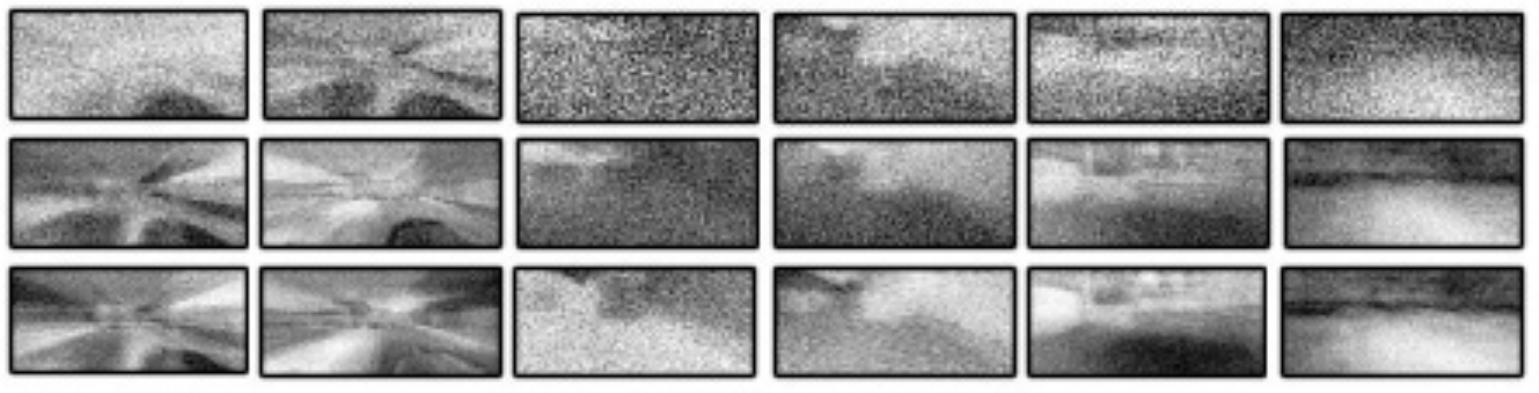}
\vspace*{-0.6cm}
\caption{Visualization of 18 filters learned by the first to third layer of a three-layered RA-DAE (top row to bottom row). The representation of structures are more visible and clear in higher layers.}
\label{fig:filters}
\vspace{-0.3cm}
\end{figure}

To understand the feature representation capabilities of distinct layers, we visualize features learned by the models using the activation maximization procedure~\cite{erhan2009visualizing}. Figure~\ref{fig:filters} visualizes the hidden layers for the three layered RA-DAE. It can be observed that the deeper the layer is, the more detailed the representation. The first layer of the network focuses on various shadows and edges, where as the third layer network represents more defined structures.

\begin{figure*}
\centering
\hspace*{0.1cm}
\includegraphics[width=0.9\textwidth,trim={2cm 0cm 0cm 0cm},clip]{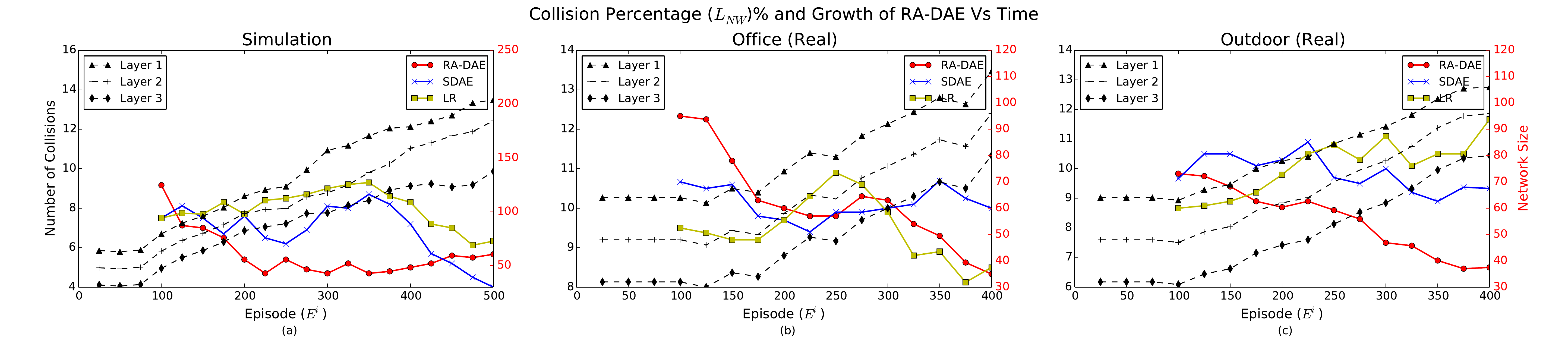}
\vspace*{-0.5cm}
\caption{Plot of the percentage of collision ($L_{NW}$) and the growth of the network over time in simulation,  office and outdoors, respectively. The left and right axes denote the percentage of collisions and the number of neurons in each layer of the RA-DAE, respectively. The simulation results indicate that in general, RA-DAE performs better than SDAE and LR in both environments. Finally, the behavior of the size of the network suggests that RA-DAE increases its complexity as the complexity of the environment increases.}
\vspace{-0.5cm}
\label{fig:bumps}
\end{figure*}
\vspace{-0.1cm}
\subsection{Comparisons}

\subsubsection{Overview}
Several experiments were performed comparing the performance of RA-DAE, SDAE and LR. For each algorithm and environment, two experiments were averaged and was taken as the result. A limit of two experiments per algorithm and environment combination was set as the algorithms displayed similar patterns in terms of the number of collisions over time. The first 100 episodes were disregarded to allow the algorithms to learn useful parameters before being compared. The number of collisions $L_{NW}$ was calculated for consecutive batches of 25 episodes, $L_{NW}=\{L_{NW}^{0:25},\ldots,L_{NW}^{N-25:N}\}$. To facilitate the interpretation of the results we converted the $L_{NW}$ values to percentages (i.e. $L_{NW}^{i-25:i}\times(100\div25)\% \hspace{0.1cm} \forall i$).

\subsubsection{Simulation Results}
The simulation results in Fig~\ref{fig:bumps}a indicate a clear reduction of $L_{NW}$ for RA-DAE, SDAE and LR over time. However, early in the learning process RA-DAE shows the steepest reduction in the number of collisions. This can be attributed to RA-DAE's ability to incrementally learn features on demand, as opposed to trying to learn with a fixed number of neurons. SDAE achieves the lowest percentage of collisions but only very late in the process. LR shows the worst performance as it is unable to deal with the complexity of the environment.

\subsubsection{Real-World Experimental Results}
\begin{figure*}[h]
\vspace{-0.2cm}
  \hspace{.1cm}
  \includegraphics[width=.92\textwidth]{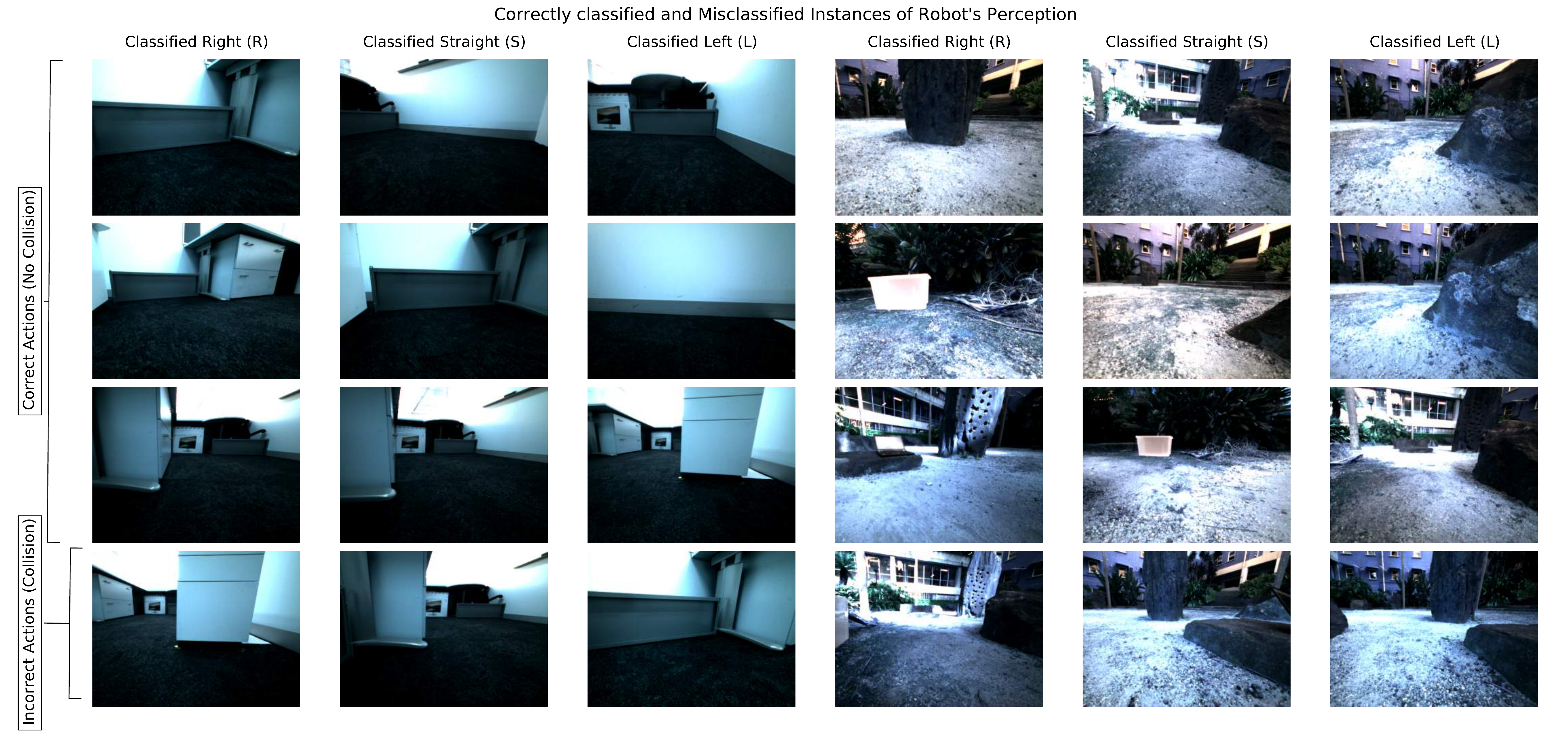}
 \vspace{-0.2cm}
\caption{The first three rows denote correctly classified instances where the last row (fourth row) illustrates misclassified instances. The first three columns illustrate samples from the office and the rest is from the outdoor environment. It is clear that the robot learns to turn depending on the obstacle position. Finally, the challenges of navigating the outdoor environment can be understood by observing the over exposures, variable lighting, etc.}
\vspace{-0.6cm}
\label{fig:classifications}
\end{figure*}

Fig~\ref{fig:bumps}b and ~\ref{fig:bumps}c show the results obtained in real-world environments; an office environment (Fig~\ref{fig:env}c) and an outdoor environment (Fig~\ref{fig:env}d). In the office environment, it can be observed that RA-DAE and LR demonstrate better performance than SDAE. The reason for LR's slightly better performance can be related to the nature of the environment; the office environment was comparatively easy to navigate as the area was small and had consistent lighting throughout the experiment, enabling LR to perform better. SDAE's slightly poor performance at the end of the experiment can be ascribed to a slight over-fitting caused by the combination of the consistency of obstacles and the complex neural structure. The outdoor environment provided a more challenging and dynamic environmental conditions such as lighting changes, making the learning more challenging. Figure~\ref{fig:classifications} shows the variability of the lighting conditions in the outdoor environment. The results from the outdoor environment suggest that LR performs the worst while SDAE shows a slight reduction on the number of collisions. RA-DAE shows the highest reduction of the number of collisions. 

\begin{figure*}[h]

\begin{subfigure}{.3\textwidth}  
  \centering
  \includegraphics[height=3cm]{images/loc_office_resized.jpg}
  \caption{Office}
  \label{fig:office_image}
\end{subfigure}
\begin{subfigure}{.30\textwidth}  
  \centering
  \includegraphics[height=3cm]{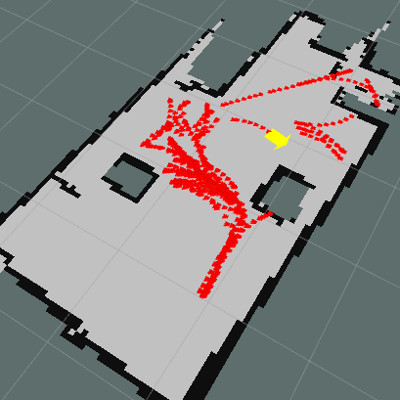}
  \caption{Traj. (Office, 50-100 Episodes)}
  \label{fig:office_traj_50_100}
\end{subfigure}
\begin{subfigure}{.30\textwidth}  
  \centering
  \includegraphics[height=3cm]{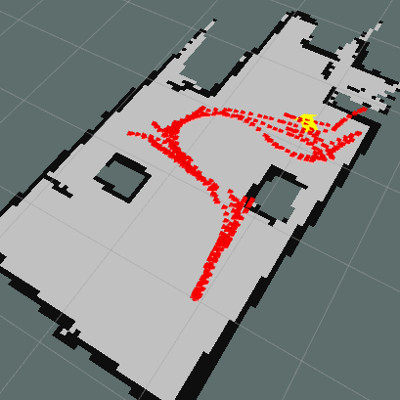}
  \caption{Traj. (Office, 350-400 Episodes)}
  \label{fig:office_traj_350_400}
\end{subfigure}
\vspace{-0.2cm}
\caption{Trajectories sampled from different time-frames in the office environment for RA-DAE. The algorithm reduces the number of collisions over time, leading to smoother trajectories.}
\label{fig:trajectories}
\end{figure*}

Moreover, the robot's ability to learn actions from images can be related to the improvement of the quality of the trajectory the robot follows (Fig~\ref{fig:trajectories}). It can be noted trajectories from episodes 350-400 are less erratic, which results in a lower number of collisions, compared to episodes 50-100. Figure~\ref{fig:classifications} shows actions selected for several images sampled from the real-world environments. In the office environment ($1^{st}, 2^{nd}$ and $3^{rd}$ row), where most obstacles are white, it can be seen that the robot goes straight if there are no white objects immediately in front of it ($2^{nd}$ row). However, when some immediate obstacle is present, the robot tends to turn right or left depending on the positioning of the obstacle. The same observation can be noted for the outdoor environment  ($4^{th}, 5^{th}$ and $6^{th}$ rows), where the robot will prefer going straight if the image contains more light areas and is not significantly occluded by a dark blob (obstacle). Also, it can be noted that it prefers to turn towards light areas when an obstacle is present. By analyzing the images classified as straight ($2^{nd}$ and $5^{th}$ rows), it can be seen that the algorithm has learned to prefer dark areas in front of it in the office environment while in the outdoor environment, the robot prefers lighter areas. Finally, in simulation, the results demonstrate that with adequate training time, SDAE can outperform RA-DAE with enough data. However, the results at the beginning of the learning process are significantly worse.

\begin{figure}
\centering
\hspace*{-0.3cm}
\includegraphics[width=0.48\textwidth]{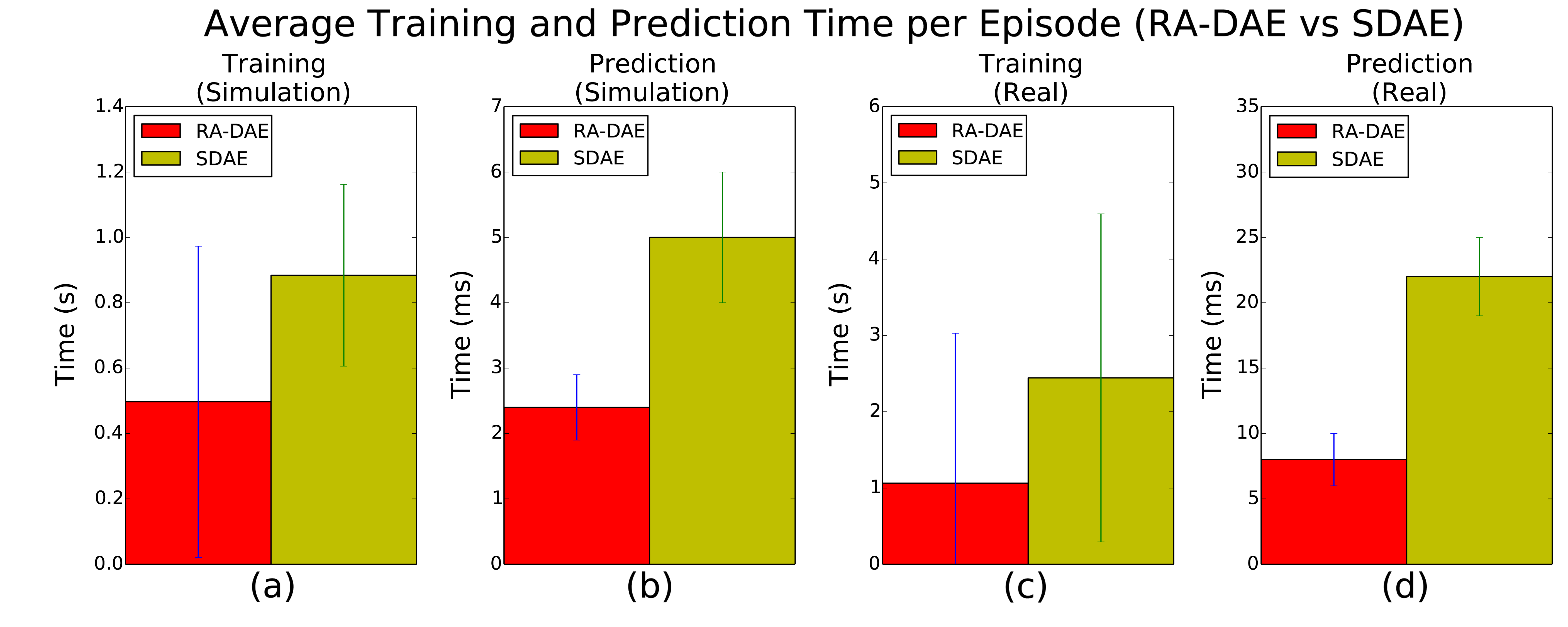}
\vspace*{-0.5cm}
\caption{Average training and prediction time for RA-DAE and SDAE. The solid bar and the error bar depict the average and standard deviation of the training and prediction time for a single episode. RA-DAE reduces the time per episode almost by half compared to SDAE.}
\label{fig:training_time}
\vspace{-0.4cm}
\end{figure}

\subsection{Evaluation of the Network Growth of RA-DAE}
The dashed lines in Figure \ref{fig:bumps} show the growth of the number of neurons in each layer of RA-DAE over time. In all experiments, RA-DAE begins with a small network and continues to grow it throughout the experiment. In the simulated environment, the network growth is aggressive compared to that in the real world experiments which can be explained by the obstacles in the simulated environment varying significantly in both color and shape, as shown in Figure \ref{fig:env}(b). RA-DAE displays a similar pattern of growth for both layers in the real world environments but less steep than in the simulated environment. Sudden drops can be noticed in the growth of the network in the office environment. Indicating the RA-DAE's attempt to learn with less neurons, as the environment is small. They are followed by increments, as the network needs more neurons to compensate for the features overridden due to continuous learning.

Using Q-learning to adapt the model structure is limited as this is effective only for discrete and small action space. For deeper networks more powerful RL techniques such as DDQN~\cite{van2016deep} should be used and will be investigated in the future.

\subsection{Evaluation of the Training Time}
Figure~\ref{fig:training_time} illustrates the training time per episode for RA-DAE and the non-adaptive equivalent SDAE. The time taken to process the data from one episode by the RA-DAE is substantially lower compared to SDAE. The reason for the faster training speed of RA-DAE results from its ability to begin with a small neural network, i.e., less parameters, and incrementally adding parameters and neurons as needed. SDAE is forced to maintain the high complexity of the network throughout the experiment resulting in longer computation time.

\section{CONCLUSIONS AND FUTURE WORK}
\vspace{-0.1cm}
Self-supervised learning remains a critical problem for autonomous navigation and long-term adaptability of robotic systems. Existing techniques rely on more expensive sensors or require extensive training sets with labels provided by humans. In this paper we developed an online self-supervised procedure based on deep neural networks that incrementally learns a predictive model, allowing a robot to navigate using a single camera. Our approach uses reinforcement learning to progressively add complexity to the network. We compare our technique (RA-DAE) to non-adaptive counterparts; Stacked Denoising Autoencoders and a Logistic Regression classifier. The experiments were conducted both in simulation and real-world environments (indoors and outdoors). The results indicate that our algorithm learns to avoid collisions comparatively better than the benchmark, while consuming less time. 

We explored online structure learning using models based on Stacked Autoencoders. An alternative model would be CNNs, or Convolutional Autoencoders. In CNNs, the filters are shared among different nodes which makes addition or merging operations more difficult. The inclusion of convolutional nets into our framework remains a topic for future work. Another avenue is the use of pre-trained models where the robot would start exploring using a well developed network, trained on another environment. The goal of RA-DAE would then be to adapt the network to a new environment.

\vspace{-0.3cm}
\bibliographystyle{named}
{\bibliography{references}}


\end{document}